\documentclass[sigconf,nonacm]{acmart}

\setcopyright{none}
\renewcommand\footnotetextcopyrightpermission[1]{}
\usepackage{booktabs}
\usepackage{amsmath}
\usepackage{array}
\usepackage{multirow}
\usepackage{graphicx}
\usepackage{balance}
\usepackage{xurl}
\usepackage{verbatim}
\usepackage{float}
\usepackage{placeins}
\usepackage{listings}
\lstdefinestyle{promptstyle}{basicstyle=\ttfamily\scriptsize,breaklines=true,breakatwhitespace=true,columns=fullflexible,keepspaces=true,xleftmargin=1em,aboveskip=2pt,belowskip=4pt}

\acmYear{2026}
\copyrightyear{2026}
\acmConference[Preprint]{arXiv preprint}{2026}{}
\acmBooktitle{}
\acmDOI{}
\acmISBN{}

\title{Point-in-Time Financial RAG with Frozen LLMs and Market-Feedback Adaptive Retrieval}

\author{Zijie Zhao}
\affiliation{%
  \institution{Massachusetts Institute of Technology}
  \city{Cambridge}
  \state{MA}
  \country{USA}}
\email{zijiezha@mit.edu}

\author{Roy E. Welsch}
\affiliation{%
  \institution{Massachusetts Institute of Technology}
  \city{Cambridge}
  \state{MA}
  \country{USA}}
\email{rwelsch@mit.edu}

\newcommand{\hset}{\ensuremath{\{1\mathrm{D},3\mathrm{D},5\mathrm{D}\}}}

\begin{document}

\begin{abstract}
Financial retrieval-augmented generation (RAG) systems typically rank evidence by textual relevance, but in financial markets evidence utility depends on event type, forecast horizon, and market context. We study news-triggered event-impact prediction as a point-in-time financial RAG problem. For each company-news anchor, the system retrieves financial news and SEC filing passages, appends a pre-decision market-context card, and predicts multi-horizon residual-return signals. Our method keeps the LLM frozen and adapts retrieval through an external Bayesian source memory updated from matured residual-return feedback. On a fixed 89-stock Nasdaq-oriented universe derived from the FinRL-DeepSeek/FNSPID task, using original FNSPID news and point-in-time EDGAR filing passages, Frozen Reader with Source Memory improves held-out macro-F1 from 0.438 to 0.471 and downstream portfolio Sharpe from 0.52 to 0.84 relative to Frozen Reader with No Memory. Supervised LoRA gives modest gains under static retrieval, but after source-memory adaptation, the LoRA reader does not improve over the frozen reader. These results suggest that, for financial RAG systems, learning where to retrieve can be as important as learning how to read, offering a modular route to market-feedback adaptation.
\end{abstract}



\maketitle

\section{Introduction}

Financial markets react to a crowded stream of public information. A single corporate event may appear in a headline, an SEC filing, an analyst note, and the recent price path, but these sources are not equally useful for judging what the market has not yet absorbed. Retrieval-augmented generation (RAG) provides a natural way to ground language models in external evidence~\cite{lewis2020rag}. In finance, however, textual relevance is only a starting point: a near-duplicate news story may restate information already priced in, while a less prominent filing passage may contain guidance, risk disclosure, or forward-looking language that better informs market impact.

We study this relevance--utility mismatch in a news-triggered setting. Instead of standalone sentiment classification or general daily return forecasting, we predict multi-horizon residual market-impact signals. Because positive news can disappoint relative to expectations and negative news can be already anticipated, we use event-study-style residual-return feedback~\cite{mackinlay1997event}: after a forecast horizon elapses, a residual-impact label indicates whether the event was followed by positive, neutral, or negative abnormal performance.

Reader-side adaptation can be useful, but realized residual-return labels are noisy and may not provide clean semantic supervision. This raises a practical question for financial RAG: can market feedback adapt evidence selection without writing these labels into reader parameters? We study this question by keeping the LLM reader fixed and adapting only the retrieval layer.

Each prediction is triggered by a timestamped company-news item from FNSPID~\cite{dong2024fnspid}, which we call a \emph{news event anchor}. The anchor defines the firm, prediction time, and event query. The system retrieves point-in-time evidence from related financial news and SEC filings, appends a pre-decision market-context card, and asks a frozen reader to output multi-horizon residual-impact predictions. Only after labels mature does an external Bayesian source memory update source-family utilities for similar event types and horizons. In this paper, ``trust'' means empirical downstream usefulness under market feedback, not factual truthfulness or causal reliability.

Market-feedback source weighting has been explored for financial sentiment analysis~\cite{zhao2024aligning}. In contrast, we study residual-return event-impact prediction with point-in-time news/SEC evidence and retrieval-side adaptation for a frozen reader. This leads to a modular framework for market-feedback adaptive financial RAG: the reader remains stable, while source-memory reranking evolves as realized residual-return feedback becomes available.

The paper makes three contributions: (i) it formulates financial RAG as news-triggered, point-in-time evidence selection for residual market-impact prediction; (ii) it introduces a frozen-reader framework in which source memory learns source-family utility by event type and forecast horizon without updating reader parameters; and (iii) it evaluates the method on original FNSPID news for a fixed 89-stock universe augmented with point-in-time EDGAR passages, showing that source-memory retrieval improves over static retrieval, reader-side LoRA offers limited additional value once source memory is active, and the resulting predictions remain useful under validity checks and a fixed downstream financial diagnostic.

\section{Related Work}

\paragraph{Financial text, event studies, and market feedback.}
Financial NLP has long studied whether textual information can predict market reactions. Recent LLM-based work shows that language-model outputs derived from news or financial text can be informative for stock-return prediction and portfolio construction~\cite{lopezlira2023chatgpt,guo2024finetuning}. Financial news datasets such as FNSPID provide aligned news and price records for studying financial time series with text~\cite{dong2024fnspid}. Event studies define abnormal returns by comparing realized returns with expected returns from a market model~\cite{mackinlay1997event}. We follow this tradition but use residual returns to construct event-level feedback labels, rather than treating them as direct sentiment labels or clean semantic annotations.

\paragraph{Financial RAG and evidence retrieval.}
Retrieval-augmented generation incorporates external context into language-model inference~\cite{lewis2020rag}. In finance, retrieved context can help interpret short headlines, firm disclosures, and long financial documents~\cite{zhang2023financialrag}. Recent financial RAG systems and benchmarks emphasize that retrieval and evidence selection are often central bottlenecks. The ACM-ICAIF FinanceRAG Challenge studied retrieval and reranking for financial documents~\cite{lee2024multireranker}; FinDER provides expert-annotated query--evidence--answer triplets for financial RAG evaluation~\cite{choi2025finder}; FinAgentBench evaluates multi-step document-type and passage retrieval in financial question answering~\cite{choi2025finagentbench}; and MultiFinRAG studies multimodal retrieval over financial reports~\cite{gondhalekar2025multifinrag}. These works mainly optimize evidence retrieval for factual question answering or answer grounding. Our work instead studies evidence selection for market-impact prediction, where evidence usefulness is measured by matured residual-return feedback rather than by static textual relevance alone. The framework is retrieval-backend flexible: source memory can be applied on top of flat lexical/dense retrieval or future graph-based retrieval, as long as retrieved evidence can be mapped to source families and timestamps~\cite{edge2024graphrag,peng2024graphrag}.

\paragraph{Reader adaptation versus retrieval-side adaptation.}
A common way to specialize LLMs for finance is to fine-tune the reader or use parameter-efficient adapters. FinGPT highlights data-centric financial LLM development and lightweight adaptation~\cite{yang2023fingpt}, while LoRA provides a general parameter-efficient mechanism for adapting large models~\cite{hu2022lora}. Recent financial studies fine-tune LLMs for sentiment or return-prediction tasks, including FinLlama for financial sentiment analysis~\cite{iacovides2024finllama}, newsflow-based stock-return prediction~\cite{guo2024finetuning}, and LoRA benchmarking on financial datasets~\cite{wang2025finlora}. These works motivate our supervised LoRA reader baseline. Our question is different: given noisy residual-return labels, can market feedback adapt the evidence-selection layer while keeping the reader frozen?

\section{Method}
\subsection{Framework Overview}

We use company news as the starting point because news provides a timestamped public event trigger and defines when the system is allowed to form a prediction. Each news item is formatted as an anchor containing the ticker, event time, and event query. The anchor is assigned to a coarse event type using the deterministic mapper in Appendix~\ref{app:event_mapping}, and the system retrieves point-in-time evidence from financial news and SEC filings. A frozen LLM reader then receives the anchor, selected evidence, and a pre-decision market-context card, and outputs one value from $\{-1,0,+1\}$ for each forecast horizon.

\begin{figure*}[t]
\centering
\includegraphics[width=0.96\textwidth]{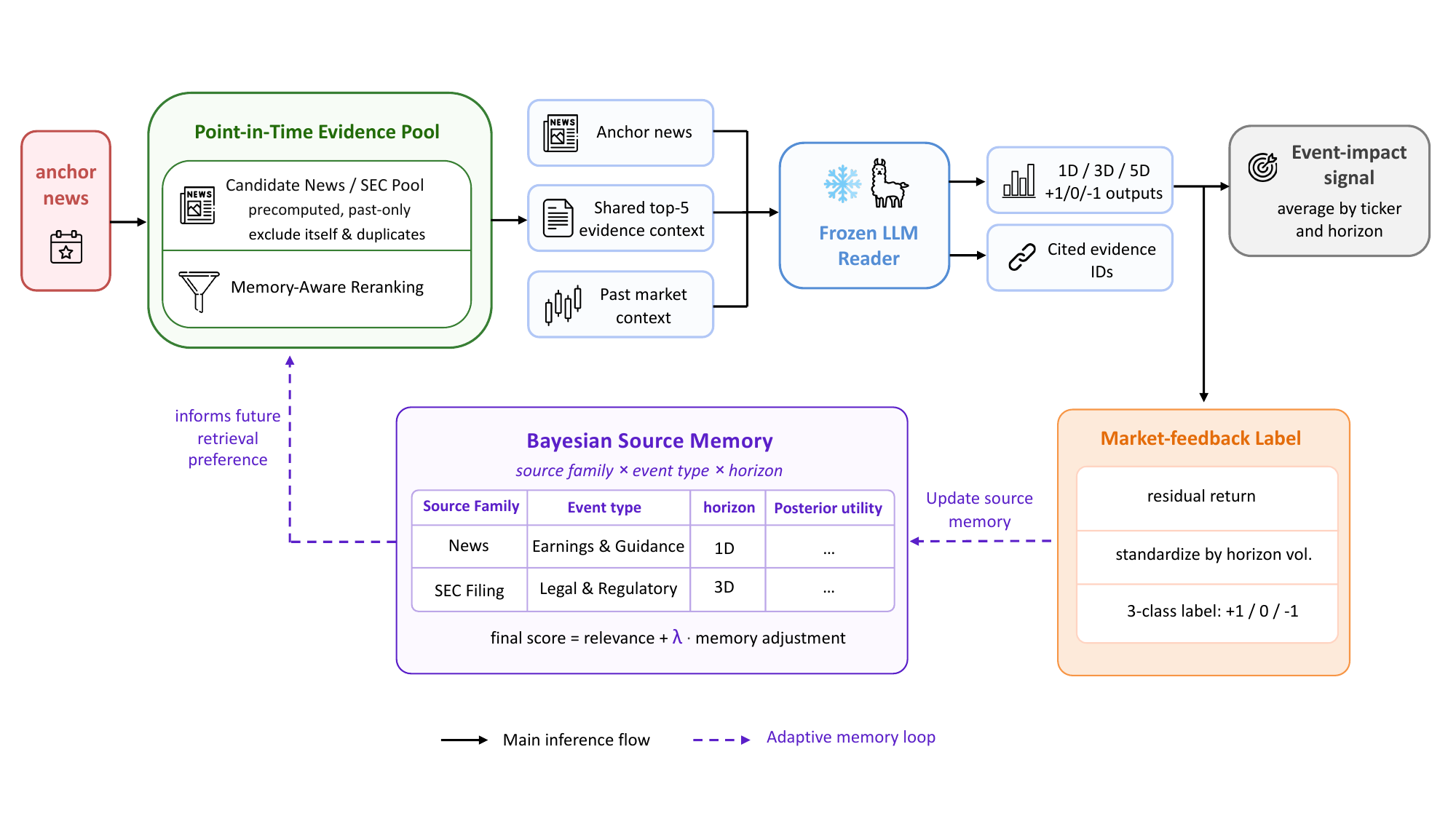}
\caption{Framework overview. News-triggered point-in-time evidence is retrieved and reranked before being read by a frozen LLM. Matured residual-return labels update an external Beta-style Bayesian source memory, which affects future retrieval only.}
\Description{A flow diagram showing news anchors, a point-in-time evidence pool, a frozen LLM reader, event-impact signals, residual-return labels, and a Bayesian source-memory feedback loop.}
\label{fig:framework}
\end{figure*}

These are anchor-level residual-impact predictions. When multiple anchors share the same ticker and prediction day, we average their predictions within each horizon to obtain a stock-day market-impact signal. After the horizon matures, the realized residual-impact label updates an external source memory, which affects future evidence ranking but not the reader parameters and does not revise the prediction that generated the feedback. Figure~\ref{fig:framework} summarizes the framework.

\subsection{Frozen LLM Reader}

We use an instruction-tuned LLM as the event-impact reader because it can synthesize heterogeneous inputs, including short news items, filing passages, and market context, into a structured multi-horizon judgment. Unless otherwise stated, the base reader is Llama-3.1-8B-Instruct, which we choose because it is an open instruction-tuned model with reproducible weights and stable structured-output behavior~\cite{meta2024llama31}. Since adaptation occurs in the retrieval layer rather than the reader parameters, the same source-memory mechanism can be paired with other open instruction readers, such as Qwen-style models~\cite{qwen2025qwen3}. The reader input consists of the event anchor, selected point-in-time evidence, the market-context card, and a fixed JSON output schema. For each forecast horizon, the reader outputs one value from $\{-1,0,+1\}$, where $+1$ denotes positive residual impact, $0$ denotes neutral or uncertain residual impact, and $-1$ denotes negative residual impact.

Reader-side adaptation can be useful, and we include a supervised LoRA reader as a controlled baseline in Section~\ref{subsec:event_quality}. The reader is frozen throughout the main method. We use a fixed prompt template, fixed JSON schema, and greedy decoding for reproducibility. Freezing is also appropriate for this task because residual-return labels are noisy market-feedback targets rather than clean semantic annotations; fitting them directly into reader parameters may encourage spurious lexical patterns. In addition, sentiment-oriented financial tuning can be misaligned with residual market-impact prediction, since positive news does not necessarily imply positive abnormal return and negative news may already be priced in. Keeping the reader fixed also makes the system easier to update when available sources or market regimes change. Appendix~\ref{app:reader_case} provides the compact prompt schema and a real-data walkthrough.

\subsection{News-Triggered Point-in-Time Evidence Preparation}

Point-in-time control is central to this task: the reader may only use evidence and market context observable at the assigned decision time. For each news-event anchor, the anchor text is included in the reader input as the event description, but it is not treated as retrieved evidence and is excluded from source-memory updates. We use a close-to-close decision convention to avoid ambiguous intraday leakage. Regular-hours anchors are assigned to a post-close decision time and are tradable from the next trading day; after-hours, weekend, and holiday anchors are assigned to the next tradable decision time. For after-close anchors, the prediction is formed after the event becomes public and before the next tradable session, using only information available by that assigned decision time; the signal enters the portfolio from the next tradable day, and all forward returns are measured after that assignment. The market-context card uses only information observable by the assigned decision time and excludes post-decision returns, future volume reactions, and future documents.

We build a global evidence store containing timestamped financial news and SEC filing passages. SEC passages are constructed from EDGAR filings matched to tickers through CIK identifiers, primarily including 8-K, 10-Q, and 10-K filings. Filings are segmented into short passages and timestamped by EDGAR acceptance datetime rather than filing date~\cite{sec2024edgar}; a filing passage is eligible only if its acceptance timestamp precedes the assigned prediction time. The anchor itself, exact URL matches, title duplicates, and duplicate-cluster matches are removed. The candidate pool is primarily company-matched. To capture cross-stock information without letting broad sector news dominate firm-specific interpretation, we allow same-sector peer evidence to occupy at most 20\% of the cached candidate pool, i.e., up to four passages when $M_0=20$.

Filtering is applied before textual scoring: future documents are excluded not only from the reader input, but also from candidate scoring and reranking. The static relevance score $r(q,d)$ is computed over the filtered candidate set using a fixed hybrid lexical--dense retriever. Specifically, we combine a BM25 lexical score and a dense cosine-similarity score from frozen sentence-embedding representations with equal weights after min--max normalizing both scores to $[0,1]$ within each anchor's candidate pool. The same filtered candidate pool, base retriever, normalization, and context budget are shared by all variants; our contribution is the market-feedback source-memory reranking layer rather than the base retriever. For offline evaluation over a fixed historical sample, we precompute at most $M_0=20$ point-in-time candidates for each anchor using only information available at that anchor's prediction time. Source memory then reranks this candidate set, and the reader receives at most $M=5$ retrieved passages.

\subsection{Market-Feedback Label Construction}

We construct market-feedback labels from post-event abnormal returns rather than human sentiment annotations. The goal is not to claim that a news item causally explains the subsequent return. Instead, the label provides residual-return feedback on whether the reader's event-impact prediction aligns with realized residual performance after controlling for broad market movement.

For each forecast horizon $h \in \hset$, let $R_h$ be the post-prediction stock return and $R^m_h$ be the market return over the same window. Returns are computed from adjusted daily prices. Because the benchmark universe is Nasdaq-oriented, we use QQQ as the market proxy. For each horizon, we estimate a simple horizon-specific market model from overlapping $h$-day stock and market returns in a rolling past-12-month window ending 20 trading days before the prediction time. The gap reduces contamination from pre-event drift and nearby news. We require at least 120 valid observations in the estimation window; otherwise the anchor-horizon pair is excluded from residual-label construction.

The post-event residual return is
\begin{equation}
\epsilon_h = R_h - (\hat{\alpha}_h + \hat{\beta}_h R^m_h).
\label{eq:residual-return}
\end{equation}
Here $\hat{\alpha}_h$ and $\hat{\beta}_h$ are estimated from pre-event $h$-day returns, so the intercept and beta are aligned with the same forecast horizon. We standardize the residual using a horizon-specific pre-event residual volatility:
\begin{equation}
z_h = \epsilon_h / \hat{\sigma}_h .
\label{eq:standardized-residual}
\end{equation}
The volatility $\hat{\sigma}_h$ is estimated from the residuals of the same pre-event horizon-specific market model. This scaling makes labels comparable across horizons.

We define the residual-impact label as
\begin{equation}
y_h =
\begin{cases}
+1, & z_h > 1,\\
0, & |z_h| \leq 1,\\
-1, & z_h < -1.
\end{cases}
\label{eq:residual-label}
\end{equation}
These labels are unavailable at prediction time and are used only after the corresponding forecast horizon has elapsed.

\subsection{Bayesian Source Memory}
\label{subsec:source_memory}

We use a lightweight Beta-style Bayesian source memory to store market-feedback utility at the source-family level rather than the document level. Each cell is indexed by source family $s$, event type $c$, and forecast horizon $h$. In our setting, the source families are financial news and SEC filings, so the memory table remains small and interpretable.

Each memory cell keeps positive and negative feedback counts, denoted by $a_{s,c,h}$ and $b_{s,c,h}$ and initialized at zero. After a forecast horizon matures, we compare the reader prediction with the realized residual-impact label. If the prediction is correct, each cited source family receives positive feedback for the corresponding $(s,c,h)$ cell; otherwise it receives negative feedback. When both source families are cited for the same horizon, credit is split equally. Cited evidence IDs are parsed from the JSON output and matched against the retrieved top-$M$ set; invalid or hallucinated IDs are ignored. If the reader does not cite a valid retrieved evidence identifier, the memory is not updated for that event-horizon pair. We update using cited evidence IDs because the goal is to assign feedback to source families indicated by the reader's cited evidence IDs, rather than to every passage placed in the context window.

Because the label $0$ represents neutral or uncertain residual impact and is the majority class in financial news data, unweighted feedback could let correct neutral predictions dominate the memory updates. We therefore use class-balanced fractional feedback: each update is multiplied by $w_y=1/(3\pi_y)$, where $\pi_y$ is the validation frequency of the realized label $y$. The factor $3$ is the number of residual-impact classes, so the expected update weight remains close to one while less frequent positive and negative residual-impact events receive proportionally larger influence.

The posterior utility of a source-memory cell is
\begin{equation}
\hat{u}_{s,c,h}
=
\frac{a_{s,c,h}+1}{a_{s,c,h}+b_{s,c,h}+2}.
\label{eq:posterior-utility}
\end{equation}
The value $0.5$ is the neutral prior utility. Values above $0.5$ indicate that a source family has historically been associated with more correct than incorrect feedback for the same event type and horizon, while values below $0.5$ indicate the opposite. A naive memory-aware reranking score would add the centered utility directly to textual relevance,
\begin{equation}
\mathrm{score}^{\mathrm{naive}}_h(q,d)
=
r(q,d) + \lambda(\hat{u}_{s(d),c,h}-0.5),
\label{eq:naive-score}
\end{equation}
where $r(q,d)$ is the normalized textual relevance score and $s(d)$ is the source family of candidate passage $d$.

This naive update has two practical issues. First, in a cold-start cell, one or two early outcomes can move the utility away from neutral even though the evidence base is weak. Second, if a source family is favored early, it can be retrieved and cited more often later, creating a policy-dependent feedback loop. We address the first issue with count-based reliability shrinkage and the second with bounded clipping. Let $n_{s,c,h}=a_{s,c,h}+b_{s,c,h}$ and $\eta_{s,c,h}=n_{s,c,h}/(n_{s,c,h}+\kappa)$. The final reranking score is
\begin{equation}
\mathrm{score}_h(q,d)=r(q,d)+\lambda\,\mathrm{clip}\!\left(\eta_{s(d),c,h}(\hat{u}_{s(d),c,h}-0.5),-\delta,\delta\right).
\label{eq:final-score}
\end{equation}
The shrinkage factor $\eta_{s,c,h}$ keeps low-count cells close to neutral, and clipping prevents source memory from overwhelming textual relevance. The coefficient $\lambda$ controls the relative strength of market-feedback source utility versus textual relevance; when $\lambda=0$, retrieval reduces to static relevance ranking. In the main experiments, we set $\kappa=30$, $\delta=0.20$, and $\lambda=0.30$ based on validation performance; Appendix~\ref{app:sensitivity} reports sensitivity to $\lambda$ and $\delta$.

For inference, the reader receives one shared context for the three horizons. We therefore compute a single reranking score by averaging the clipped memory adjustments for the 1D, 3D, and 5D cells, then multiplying that average by $\lambda$ and adding it to the static relevance score. This shared score selects the fixed top-$M=5$ context; feedback counts and posterior utilities remain horizon-specific after labels mature.

\section{Evaluation}
\subsection{Dataset, Splits, and Evaluation Protocol}

\begin{table*}[t]
\centering
\small
\setlength{\tabcolsep}{4pt}
\caption{Constructed benchmark audit and split usage. SEC cov. reports the fraction of anchors whose cached top-$M_0=20$ point-in-time candidate pool contains at least one eligible SEC filing passage. Label columns report the proportions of $+1$, $0$, and $-1$ residual-impact labels.}
\label{tab:dataset-audit}
\begin{tabular}{llrrrrccc}
\toprule
Split & Years & \#Anchors & Anchors/day & Active ticker-days & SEC cov. & 1D +1/0/-1 & 3D +1/0/-1 & 5D +1/0/-1 \\
\midrule
Train (LoRA) & 2019 & 31,270 & 124.1 & 15,120 & 33.1\% & 17.2/65.6/17.2 & 18.5/63.1/18.4 & 20.1/59.7/20.2 \\
Validation & 2020 & 36,920 & 146.0 & 17,460 & 34.7\% & 18.4/63.7/17.9 & 19.2/61.5/19.3 & 20.7/58.6/20.7 \\
Test & 2021--2023 & 112,640 & 149.6 & 54,120 & 35.4\% & 17.8/64.8/17.4 & 18.6/62.7/18.7 & 20.4/59.0/20.6 \\
\bottomrule
\end{tabular}
\end{table*}

\begin{table*}[t]
\centering
\small
\setlength{\tabcolsep}{4pt}
\caption{Event-level prediction quality on held-out test events. Non-neutral F1 averages F1 over the $+1$ and $-1$ classes only. Macro-F1 averages over all three classes and the 1D/3D/5D horizons. Rank IC is the daily cross-sectional Spearman correlation between ticker-day predictions and standardized residual returns $z_h$, averaged over days and horizons. LoRA rows report mean $\pm$ standard deviation over five training seeds.}
\label{tab:event-results}
\begin{tabular}{lcccccc}
\toprule
Method & Non-neutral F1 & Macro-F1 & 1D Macro-F1 & 3D Macro-F1 & 5D Macro-F1 & Rank IC \\
\midrule
Frozen + No Memory & .311 & .438 & .425 & .441 & .449 & .046 \\
Frozen + Source Memory & .356 & .471 & .452 & .476 & .486 & .061 \\
LoRA + No Memory & .326$\pm$.017 & .452$\pm$.014 & .438$\pm$.014 & .454$\pm$.015 & .464$\pm$.014 & .052$\pm$.006 \\
LoRA + Source Memory & .349$\pm$.016 & .468$\pm$.013 & .448$\pm$.013 & .473$\pm$.013 & .483$\pm$.012 & .058$\pm$.006 \\
\bottomrule
\end{tabular}
\end{table*}

We use the original timestamped FNSPID news records and adjusted stock-price histories because they are suitable for event-driven market-feedback evaluation. FNSPID contains 29.7 million stock-price records and 15.7 million time-aligned financial-news records for 4,775 companies from 1999 to 2023~\cite{dong2024fnspid}. For a controlled benchmark, we use the FinRL-DeepSeek/FNSPID Stock Trading task only to define a fixed 89-stock Nasdaq-oriented universe~\cite{finrl2025task1}. We do not use the reduced contest trading file, its one-news-per-stock-day sampling, or its LLM-generated sentiment and risk scores. Instead, for each ticker in the 89-stock universe, we return to the original FNSPID news stream and keep all valid timestamped news anchors after duplicate removal and point-in-time filtering. We then augment the news evidence pool with point-in-time EDGAR filing passages matched to the same ticker universe.

We construct three time-ordered splits: 2019 for LoRA reader adaptation, 2020 for validation, and 2021--2023 for held-out testing. The validation split is used for context-budget and source-memory hyperparameter choices and to fix class-balanced feedback weights. For held-out testing, we use an online prequential protocol: all source-memory feedback counts are initialized to zero at the beginning of 2021, and the memory is updated only with feedback from earlier test-period anchors whose forecast horizons have already matured. Validation labels are not used to pretrain or warm-start the test-period memory. Table~\ref{tab:dataset-audit} summarizes the resulting event-level benchmark. The class associated with label $0$ is the largest class, which is expected because many news events are not followed by statistically large residual moves. The positive and negative classes are smaller than the neutral class but remain large enough for non-neutral F1 evaluation.

\subsection{Event-Level Prediction Quality}
\label{subsec:event_quality}
We compare four variants formed by two design choices: whether the reader is kept frozen or adapted with LoRA, and whether retrieval is static or adjusted by source memory. In the No Memory variants, the system still uses the same point-in-time RAG evidence and fixed reader context; only the Bayesian source-memory reranking term is disabled. For reader-side adaptation, we train LoRA adapters on the Train split using matured residual-impact labels while keeping the base LLM weights fixed. The LoRA reader uses the same backbone, prompt, and context format as the frozen reader, but its supervised loss is applied only to horizon-level residual-impact labels; evidence IDs and rationales are produced at inference time and are not used as supervised citation targets. We use rank 16, alpha 32, dropout 0.05, learning rate $2\times10^{-4}$, effective batch size 64, adapters on attention projection layers, and validation Macro-F1 for early stopping and checkpoint selection. This setup gives a controlled comparison between adapting the reader and adapting the retrieval layer.

Table~\ref{tab:event-results} reports event-level prediction quality on the held-out test split. Besides all-class macro-F1, we report Non-neutral F1, which excludes the label $0$ and evaluates performance on positive and negative residual-impact events. This metric tests whether gains come from positive and negative market-impact events rather than only from the majority neutral class. With the reader frozen, source memory improves Non-neutral F1 from .311 to .356, Macro-F1 from .438 to .471, and Rank IC from .046 to .061. LoRA adaptation improves the static-retrieval reader, but the gain is smaller than the gain from source-memory retrieval. When source memory is active, the LoRA-adapted reader remains close to but does not improve over the frozen reader. On this benchmark, source-memory retrieval yields a larger event-level gain than reader-side LoRA adaptation.

This pattern is consistent with the design choice of using noisy residual-return feedback to adapt retrieval preferences rather than fitting the reader directly to market labels. The improvement from source memory is also consistent with the relevance--utility mismatch in financial RAG. Textual similarity does not always imply market usefulness: near-duplicate news may repeat information already reflected in prices, while filings may contain guidance or risk language that matters over longer horizons. A source-family memory indexed by event type and horizon helps the retriever prefer sources that have historically been useful for similar market-feedback decisions.

\subsection{Source-Memory Validity Checks}

We isolate the source-memory layer under the frozen-reader design to test whether gains come from the intended feedback structure rather than incidental reranking. No memory is the static-retrieval baseline, and source memory is the main variant. The top-5 all-source update uses all source families in the final top-5 context instead of cited IDs; its weaker result suggests that cited IDs provide more targeted credit assignment. Shuffling feedback labels brings performance close to no memory, while shuffling event types weakens but does not remove the gain, suggesting both global and event-specific source utility.

We also audit selected evidence: the future-document rate is zero, near-duplicate anchor matches stay below 1\%, and valid cited-ID rates remain around 88\%. These checks suggest that the gains do not come from future evidence, anchor self-retrieval, or loss of context use.

\begin{table}[t]
\centering
\footnotesize
\setlength{\tabcolsep}{3.4pt}
\caption{Source-memory validity checks. All rows use the frozen reader; ``+'' rows modify the memory update rule or key. Shuffled rows report mean $\pm$ std over five shuffles.}
\label{tab:memory-validity}
\resizebox{\columnwidth}{!}{%
\begin{tabular}{@{}lccc@{}}
\toprule
Variant & Non-neut. F1 & Macro-F1 & Rank IC \\
\midrule
No Memory & .311 & .438 & .046 \\
Source Memory & .356 & .471 & .061 \\
+ Top-5 all-source update\textsuperscript{*} & .348 & .466 & .058 \\
+ Shuffled feedback labels & .316$\pm$.004 & .441$\pm$.003 & .047$\pm$.002 \\
+ Shuffled event types & .336$\pm$.005 & .456$\pm$.004 & .054$\pm$.003 \\
\bottomrule
\end{tabular}%
}
\begin{flushleft}
\footnotesize \textsuperscript{*}Top-5 all-source update ignores cited evidence IDs and assigns feedback to every distinct source family appearing in the final top-5 context.
\end{flushleft}
\end{table}

\begin{figure*}[t]
  \centering
  \includegraphics[width=0.88\textwidth]{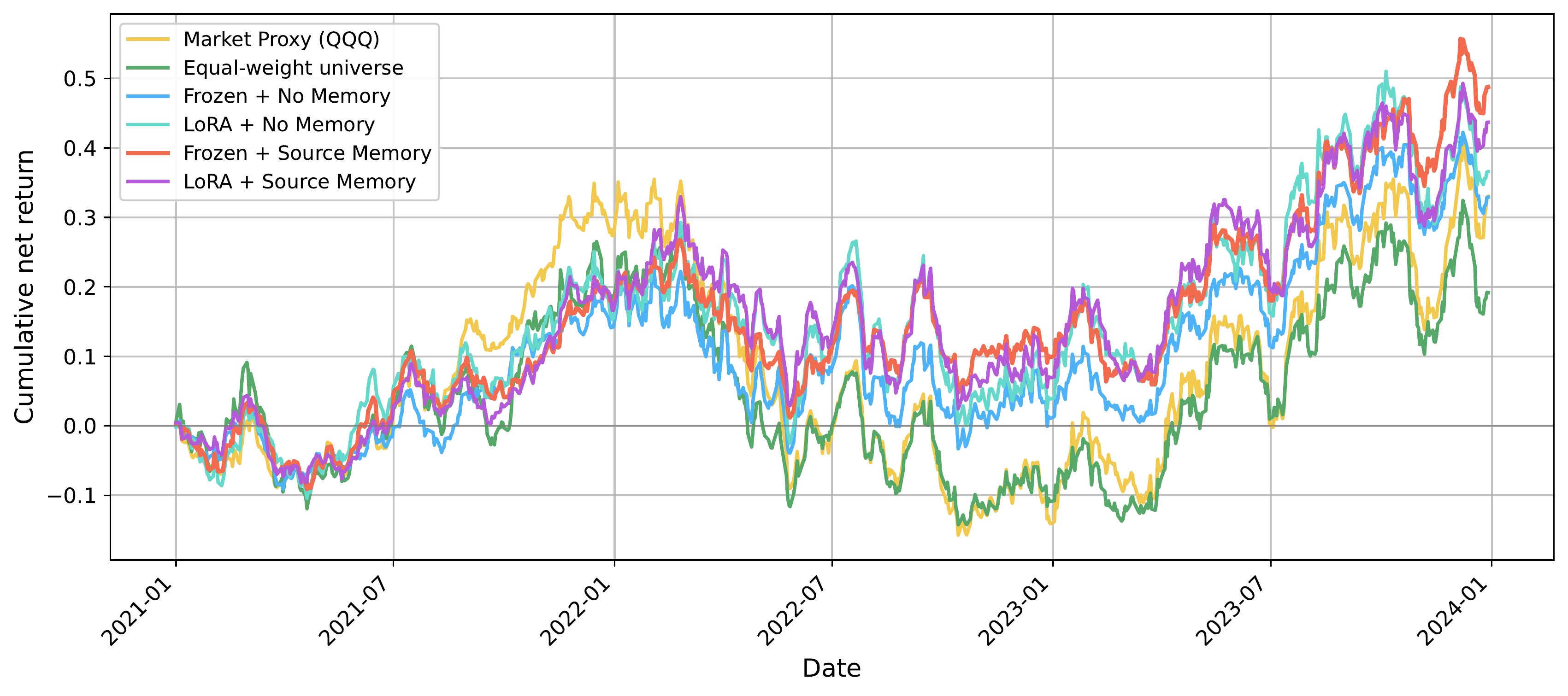}
  \caption{Daily cumulative net returns for the fixed 10 bps cost-adjusted downstream portfolio diagnostic over the held-out test period.}
  \Description{A cumulative return line chart over 2021 to 2023 comparing QQQ, equal-weight universe, Frozen and LoRA variants with and without source memory. The Frozen plus Source Memory line ends highest.}
  \label{fig:portfolio-curve}
\end{figure*}

\subsection{Downstream Portfolio Diagnostic}

We use a fixed portfolio diagnostic as a downstream financial decision task rather than an optimized trading strategy. The diagnostic rule is fixed before held-out evaluation and is not tuned on test-period returns. Its input is the 3D stock-day signal from each model. On each decision day, stocks with positive 3D scores are eligible for selection; ties are broken deterministically by the average static retrieval score and then by ticker symbol. We select up to $K=10$ names per sleeve and assign equal weights. If fewer than $K=10$ names are selected, the remaining sleeve weight is held in cash; if no names are selected, the sleeve remains in cash.

Each selected sleeve is held for three trading days, so the implemented portfolio combines overlapping sleeves. Let $v_t$ denote the equal-weight sleeve selected using the decision-day signal at time $t$. The implemented portfolio on day $t$ is
\begin{equation}
w_t = \frac{1}{3}(v_{t-1}+v_{t-2}+v_{t-3}).
\label{eq:portfolio-weight}
\end{equation}
Daily turnover is measured as one-way portfolio turnover,
\begin{equation}
TO_t=\frac{1}{2}\sum_i |w_{i,t}-\tilde{w}_{i,t}|,
\label{eq:turnover}
\end{equation}
where $\tilde{w}_{i,t}$ denotes the pre-rebalance portfolio weight after price drift. Net returns are
\begin{equation}
r^{\mathrm{net}}_t=r^{\mathrm{gross}}_t-0.001\,TO_t,
\label{eq:net-return}
\end{equation}
corresponding to a 10 bps proportional cost proxy applied to traded turnover.

All portfolio metrics are computed from daily net returns over the held-out test window. Annualized return is computed as 252 times the mean daily net return. Annualized volatility is $\sqrt{252}$ times the standard deviation of daily net returns, and Sharpe ratios use a zero risk-free rate with $\sqrt{252}$ annualization. The equal-weight universe benchmark uses the same fixed ticker set and adjusted daily prices, is rebalanced daily to equal weights, and is charged the same turnover-cost rule. The QQQ benchmark is buy-and-hold over the same test window and therefore has zero portfolio turnover in Table~\ref{tab:portfolio-results}.

\begin{table}[t]
\centering
\scriptsize
\setlength{\tabcolsep}{2.2pt}
\caption{Portfolio diagnostic from 3D stock-day predictions. Returns are net of a 10 bps turnover cost.}
\label{tab:portfolio-results}
\resizebox{\columnwidth}{!}{
\begin{tabular}{lccccc}
\toprule
Strategy & Ann. Ret. & Ann. Vol. & Sharpe & Max DD & Turnover \\
\midrule
Market Proxy (QQQ) & 10.0\% & 24.1\% & .41 & -37.8\% & .000 \\
Equal-weight universe & 6.0\% & 23.7\% & .25 & -33.5\% & .004 \\
Frozen + No Memory & 9.9\% & 19.1\% & .52 & -21.5\% & .094 \\
LoRA + No Memory & 10.9\% & 23.1\% & .47 & -24.6\% & .101 \\
Frozen + Source Memory & 14.2\% & 16.9\% & .84 & -20.2\% & .089 \\
LoRA + Source Memory & 12.8\% & 19.2\% & .67 & -22.9\% & .096 \\
\bottomrule
\end{tabular}
}
\end{table}

Table~\ref{tab:portfolio-results} shows that source-memory retrieval gives the strongest return-risk trade-off under this fixed rule: Frozen + Source Memory raises Sharpe from .52 to .84 relative to Frozen + No Memory while slightly reducing turnover. The LoRA variants do not exceed this source-memory frozen-reader result on Sharpe or drawdown. Figure~\ref{fig:portfolio-curve} plots cumulative net returns for the same fixed diagnostic. These results do not establish an optimal trading system, but show that the framework produces event-impact signals that remain useful under a fixed, reproducible, cost-adjusted downstream financial decision rule, serving as a proof of concept for a practical financial RAG workflow.

\section{Discussion and Conclusion}

Source-memory retrieval improves event-level prediction quality and downstream financial diagnostics. With the reader frozen, source memory improves non-neutral F1, macro-F1, and rank IC relative to static retrieval, and the fixed portfolio diagnostic shows a stronger return-risk trade-off under the same rule and transaction-cost assumption. The shuffled-label, shuffled-event-type, and top-5 all-source checks indicate that the gains are not explained by random reranking and that cited-evidence feedback is important to the memory mechanism. Reader-side LoRA provides modest gains when retrieval is static, but once source memory is active, the LoRA reader remains close to and does not improve over the frozen reader. These results suggest that, in this setting, adapting evidence selection is more useful than fitting noisy residual-return labels directly into reader parameters.

Realized residual-return feedback is noisy and may reflect macro shocks, liquidity, sector rotation, or concurrent firm events; source utility should therefore be interpreted as historical downstream usefulness, not causal attribution or source truthfulness. Source-memory cells are sparse at the beginning, and early retrieval choices can influence which sources later receive feedback. Shrinkage, clipping, and shuffled-memory checks mitigate cold-start and self-reinforcement effects but do not fully remove them. Finally, the evaluation uses a controlled Nasdaq-oriented universe and two broad source families; the fixed 89-stock universe may introduce universe selection or survivorship effects, so portfolio results should be interpreted only as relative diagnostics under the same fixed rule. The portfolio analysis is a fixed downstream diagnostic and does not imply future profitability. Any real deployment would require human oversight, compliance review, and risk controls.

This points to a practical design for financial RAG systems: keep the LLM stable and evidence-grounded, and adapt the retrieval layer as market regimes and information sources evolve. This separation is useful because the reader can remain reproducible and auditable, while the external source memory provides a lightweight and interpretable mechanism for incorporating delayed market feedback. More broadly, market-feedback adaptive retrieval offers a modular route to financial RAG systems that learn not only what text is relevant, but which sources have historically been useful for a given event type, horizon, and decision context.

\bibliographystyle{ACM-Reference-Format}
\bibliography{paper}
\appendix

\section{Benchmark and Implementation Details}
\label{app:implementation}

\subsection{Event-Type Mapping}
\label{app:event_mapping}

Event types are assigned before retrieval by a deterministic rule-based mapper over the news anchor. The mapper uses only anchor text and metadata, not retrieved evidence, market feedback, or future returns.

For each anchor, we lowercase the headline and available summary text and count matched keyword or phrase cues for each event type in Table~\ref{tab:eventtype}. The anchor is assigned to the event type with the largest cue count. If no event type has a matched cue, or if the top count is tied across multiple event types, the anchor is assigned to Other/Mixed. The counting rule avoids fixed-priority decisions when a news item contains multiple event cues.

\begin{table}[H]
\centering
\scriptsize
\setlength{\tabcolsep}{3pt}
\caption{Keyword and phrase cues for deterministic event-type assignment.}
\label{tab:eventtype}
\resizebox{\columnwidth}{!}{%
\begin{tabular}{lp{0.70\columnwidth}}
\toprule
Event type & Keyword or phrase cues \\
\midrule
Earnings \& Guidance &
earnings, EPS, revenue, sales, quarter, quarterly results, fiscal Q, beat, miss, guidance, outlook, forecast, raises guidance, cuts guidance, margin, EBITDA, profit, loss \\

Legal \& Regulatory &
lawsuit, litigation, class action, investigation, probe, subpoena, settlement, fine, penalty, enforcement, antitrust, DOJ, FTC, FDA, FDA approval, regulatory approval, recall, compliance, consent order \\

Capital \& Transactions &
merger, acquisition, M\&A, takeover, buyout, deal, divestiture, spin-off, IPO, offering, secondary offering, debt, notes, refinancing, credit facility, buyback, repurchase, dividend, stake, investment \\

Management \& Operations &
CEO, CFO, COO, board, appoints, resigns, departure, management change, partnership, contract, customer win, product launch, supply chain, facility, plant, production, layoffs, restructuring, workforce \\

Other/Mixed &
no matched cue, or tied top cue counts across multiple event types \\
\bottomrule
\end{tabular}}
\end{table}

\subsection{Point-in-Time Assignment and Residual Labels}
Regular-hours anchors are assigned to a post-close decision time and are tradable from the next trading day; after-hours, weekend, and holiday anchors are assigned to the next tradable decision time. For after-close anchors, the prediction is formed after the event becomes public and before the next tradable session, using only information available by that assigned decision time. The signal enters the portfolio from the next tradable day, and all forward returns are measured after that assignment.

For residual labels, we estimate the horizon-specific market model from overlapping $h$-day stock and market returns in a rolling past-12-month window ending 20 trading days before the prediction time. The gap reduces contamination from pre-event drift and nearby news. We require at least 120 valid observations in the estimation window; otherwise the anchor-horizon pair is excluded.

\subsection{Reader Adaptation and Retrieval Details}
The LoRA reader uses rank 16, alpha 32, dropout 0.05, learning rate $2\times10^{-4}$, effective batch size 64, and adapters on attention projection layers. Checkpoints are selected by validation Macro-F1. The static relevance score combines BM25 lexical retrieval and dense cosine similarity from frozen sentence-embedding representations with equal weights after min--max normalizing both scores to $[0,1]$ within each anchor's candidate pool.

\section{Real-Data Case Study and Prompt Schema}
\label{app:reader_case}

This appendix gives a real AAPL earnings example to show the anchor format, point-in-time evidence, market-context card, prompt schema, and memory update.

We use an AAPL earnings-related news anchor dated April 30, 2020. The anchor headline is \emph{Apple Reports Big Q2 Earnings Beat, Driven By Record Services Revenue}, corresponding to a Benzinga earnings article reporting Apple's fiscal Q2 2020 results~\cite{benzinga2020apple}. Apple also furnished a Form 8-K on April 30, 2020. The filing reports fiscal Q2 results under Item 2.02, ``Results of Operations and Financial Condition,'' and attaches the corresponding earnings press release as Exhibit 99.1~\cite{sec2020apple8k}. Apple press materials report revenue of \$58.3B, diluted EPS of \$2.55, Services revenue of \$13.3B, a quarterly dividend of \$0.82 per share, and an additional \$50B share-repurchase authorization~\cite{apple2020q2}.

The anchor is formatted as follows:
\begin{lstlisting}[style=promptstyle]
A0:
ticker: AAPL
company: Apple Inc.
anchor_time: 2020-04-30 after close
prediction_day: 2020-05-01
headline: Apple Reports Big Q2 Earnings Beat,
          Driven By Record Services Revenue
event_type: Earnings & Guidance
\end{lstlisting}

The anchor is always included in the reader input as the event description. It is not treated as retrieved evidence, is removed from the retrieval candidate pool, and is not used for source-memory updates. For offline evaluation, the system first constructs a point-in-time candidate pool of at most $M_0=20$ passages using only documents observable before the prediction time. The pool is primarily company-matched, with at most 20\% reserved for same-sector peer evidence. After source-memory reranking with averaged horizon adjustments, the reader receives the fixed top-$M=5$ shared context. A compact excerpt of the selected context for this event is:
\begin{lstlisting}[style=promptstyle]
N1 [News]:
Apple CEO Tim Cook says the company will not issue
guidance for the following quarter.

F1 [SEC Filing]:
Apple Form 8-K, Item 2.02, dated April 30, 2020,
reporting results of operations and
financial condition.

F2 [SEC Filing]:
Exhibit 99.1 press release attached to the 8-K,
reporting revenue, diluted EPS, Services revenue,
dividend increase, and share repurchase
authorization.
\end{lstlisting}

The market-context card is generated deterministically from adjusted daily prices available no later than the assigned decision close. It summarizes recent stock returns, QQQ-relative performance, recent realized volatility, and the recent market trend using fixed threshold rules chosen on the validation split and then held fixed for the test period. The card is appended to the reader input but is not a retrieved evidence source, is not indexed by source memory, and does not receive source-memory feedback. For this example, the event is treated as tradable on May 1, 2020, so the card uses market data available through the April 30, 2020 close:
\begin{lstlisting}[style=promptstyle]
Market context card:
as_of: 2020-04-30 close
market_proxy: QQQ

AAPL recent return:
prior 1D:  +2.10%
prior 5D:  +6.82%
prior 20D: strong positive momentum

relative performance:
AAPL outperformed QQQ over the short
pre-event window

risk state:
elevated realized volatility for both AAPL and QQQ

market trend:
positive rebound regime with elevated volatility
\end{lstlisting}
The card does not contain post-event returns, residual-impact labels, future volume reactions, or future documents.

\noindent The reader receives the anchor, selected evidence, market-context card, and a fixed output schema. A compact prompt is:
\begin{lstlisting}[style=promptstyle]
SYSTEM:
You are a financial event-impact reader.
Use only the
provided event anchor, point-in-time evidence, and
pre-decision market context. Return valid JSON only.

USER:
Predict residual market impact for fixed horizons:
1D, 3D, and 5D.
Allowed labels: +1 positive, 0 neutral/uncertain,
-1 negative.

Event anchor:
A0: Apple Reports Big Q2 Earnings Beat, Driven By
Record Services Revenue

ticker: AAPL
prediction_day: 2020-05-01
event_type: Earnings & Guidance

Retrieved evidence:
N1: Apple will not issue guidance next quarter.
F1: Apple Form 8-K Item 2.02 reporting
fiscal Q2 results.
F2: Exhibit 99.1 press release with revenue, EPS,
Services, dividend, and buyback information.

Market context:
AAPL had strong recent positive momentum,
outperformed
QQQ over the short pre-event window,
and traded in an
elevated-volatility market rebound.
\end{lstlisting}

The required JSON output schema is:
\begin{lstlisting}[style=promptstyle]
{
  "event_type": string,
  "signals": {
    "1D": {
      "signal": "-1|0|+1",
      "ids": [retrieved_evidence_id],
      "reason": string
    },
    "3D": {
      "signal": "-1|0|+1",
      "ids": [retrieved_evidence_id],
      "reason": string
    },
    "5D": {
      "signal": "-1|0|+1",
      "ids": [retrieved_evidence_id],
      "reason": string
    }
  },
  "main_uncertainty": string
}
\end{lstlisting}

After each forecast horizon matures, the reader prediction is compared with the residual-impact label from the market-feedback pipeline. For example, if the 3D prediction is correct and cites one SEC filing passage, the SEC/Earnings/3D memory cell receives $w_y$ positive feedback. If both News and SEC Filing are cited for the same horizon, each source family receives $0.5w_y$. Incorrect predictions update the corresponding negative-feedback count in the same way. The class-balanced factor $w_y$ is defined in Section~\ref{subsec:source_memory}. The update changes future retrieval preferences for similar event-type and horizon combinations; it does not change the frozen reader parameters or revise the prediction that generated the feedback.

\section{Source-Memory Strength Sensitivity}
\label{app:sensitivity}

We tune the retrieval-strength coefficient $\lambda$ and clipping bound $\delta$ on the validation split and fix them for held-out evaluation. The coefficient $\lambda$ controls the relative weight of source memory versus textual relevance, while $\delta$ limits the maximum source-memory adjustment. Table~\ref{tab:memory-sensitivity} shows that moderate memory strength improves event-level and portfolio metrics, while overly large values reduce performance by allowing source memory to override textual relevance.

\begin{table}[H]
\centering
\scriptsize
\setlength{\tabcolsep}{4pt}
\caption{Validation sensitivity to source-memory strength. The selected setting is $\lambda=0.30$ and $\delta=0.20$.}
\label{tab:memory-sensitivity}
\resizebox{\columnwidth}{!}{%
\begin{tabular}{llcccc}
\toprule
Setting & Value & Non-neutral F1 & Macro-F1 & Rank IC & Sharpe \\
\midrule
\multicolumn{6}{l}{\emph{Varying $\lambda$ with $\delta=0.20$ fixed}} \\
$\lambda$ & 0.00 & .309 & .436 & .044 & .50 \\
$\lambda$ & 0.10 & .331 & .451 & .053 & .64 \\
$\lambda$ & 0.20 & .341 & .464 & .057 & .76 \\
$\lambda$ & 0.30 & .352 & .468 & .060 & .80 \\
$\lambda$ & 0.50 & .347 & .462 & .056 & .74 \\
\midrule
\multicolumn{6}{l}{\emph{Varying $\delta$ with $\lambda=0.30$ fixed}} \\
$\delta$ & 0.05 & .329 & .451 & .052 & .63 \\
$\delta$ & 0.10 & .344 & .461 & .057 & .74 \\
$\delta$ & 0.15 & .348 & .469 & .061 & .78 \\
$\delta$ & 0.20 & .352 & .467 & .060 & .80 \\
$\delta$ & 0.30 & .345 & .463 & .056 & .71 \\
\bottomrule
\end{tabular}}
\end{table}

We select $\lambda=0.30$ and $\delta=0.20$ because this setting gives the best validation Non-neutral F1 and Sharpe while remaining near the best Macro-F1 and Rank IC. The maximum memory contribution is then $\lambda\delta=0.06$, enough to rerank relevance-comparable candidates while keeping textual relevance primary.

\end{document}